\documentclass[10pt,twocolumn,letterpaper]{article}

\usepackage{cvpr}              

\usepackage{graphicx}
\usepackage{amsmath}
\usepackage{amssymb}
\usepackage{booktabs}
\usepackage{enumitem}

\usepackage[pagebackref,breaklinks,colorlinks]{hyperref}

\usepackage[capitalize]{cleveref}
\crefname{section}{Sec.}{Secs.}
\Crefname{section}{Section}{Sections}
\Crefname{table}{Table}{Tables}
\crefname{table}{Tab.}{Tabs.}

\def\cvprPaperID{*****} 
\def\confName{CVPR}
\def\confYear{2022}

\begin{document}

\title{Multi-task 3D building understanding with multi-modal pretraining}

\author{Shicheng (Luke) Xu\\
Stanford University (SCPD)\\
\& Google \\
{\tt\small shicheng@google.com}}
\maketitle

\begin{abstract}
   This paper explores various learning strategies for 3D building type classification and part segmentation on the BuildingNet dataset\cite{selvaraju2021buildingnet}. ULIP with PointNeXt\cite{xue2022ulip} and PointNeXt segmentation\cite{qian2022pointnext} are extended for the classification and segmentation task on BuildingNet dataset.
   The best multi-task PointNeXt-s model with multi-modal pretraining achieves 59.36 overall accuracy for 3D building type classification, and 31.68 PartIoU for 3D building part segmentation on validation split. The final PointNeXt-XL model achieves 31.33 PartIoU and 22.78 ShapeIoU on test split for BuildingNet-Points segmentation, which significantly improved over PointNet++ model reported from BuildingNet paper, and it \textbf{won the 1st place} in the BuildingNet challenge at CVPR23 StruCo3D workshop\footnote{\url{https://eval.ai/web/challenges/challenge-page/1938/leaderboard/4590}}.
\end{abstract}

\section{Introduction}
\label{sec:Introduction}

Understanding building types and building parts in 3D has wide applications in mapping, autonomous driving, architecture, construction, and gaming. Buildings are dominant objects in urban areas and are what most humans interact with daily. While architecture has a long history and building follows some ontology and style, the digitization process and understanding building semantics in 3d have been extremely challenging.

The BuildingNet dataset\cite{selvaraju2021buildingnet} is a public research dataset for building exteriors and surroundings with segmentation of building parts. It includes about 2k 3d buildings of office, school, castle, house, hotel, etc; with 31 common semantic parts, such as wall, window, roof, floor, stairs, etc. There are several indoor 3D scene datasets, such as SceneNet\cite{handa2015scenenet}, S3DIS\cite{armeni_cvpr16}, and CVPR23 CV4AEC workshop \footnote{\url{https://cv4aec.github.io/}}; as well as outdoor 3D scene datasets, such as Waymo open dataset\cite{sun2020scalability}, and KITTI-360\cite{liao2022kitti}, that also annotates buildings, but none of them annotate building exterior parts. In parallel, researchers have also developed several methods to create synthetic buildings \cite{fedorova2021synthetic}, \cite{tono2022vitruvio}.

In this paper, we will look at both 3D building type classification and part segmentation on the BuildingNet dataset. For 3D building type classification, the model needs to produce a class label for the entire building object, e.g. house, hotel, etc. As for 3D building part segmentation, the model needs to assign a part label for every point in the 3D point cloud, e.g. roof, window, etc. Comprehensive studies on various transfer learning strategies will show what multi-task training and multi-modal pretraining are effective approaches to reduce overfitting and improve performance on both tasks. The best multi-task PointNeXt-s model with multi-modal pretraining achieves 59.36 overall accuracy for 3D building type classification, and 31.68 PartIoU for 3D building part segmentation on validation split.

\begin{figure}[t]
\centering
\includegraphics[width=0.8\linewidth]
               {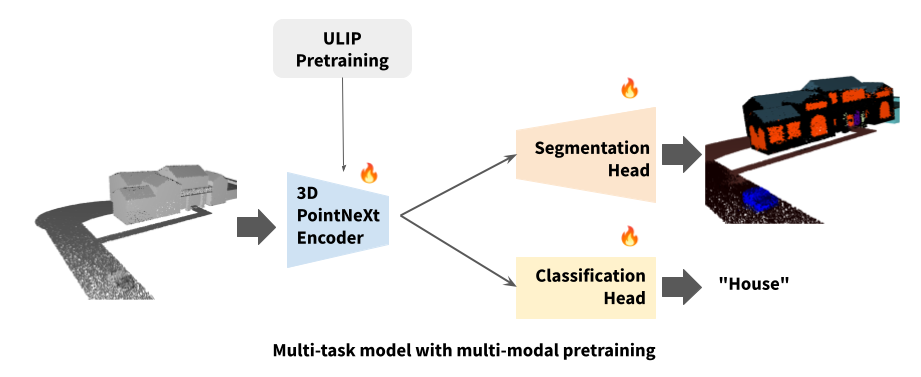}
\caption{Proposed model architecture for 3D building classification and segmentation.}
\label{fig:model_architecture}
\end{figure}

The CVPR23 StruCo3D workshop hosts the BuildingNet challenge on building part segmentation task, with an updated BuildingNet v1 dataset with two phases. This project focuses on the Building-Points phase, which is designed for large-scale point-based processing algorithms that must deal with unstrctured point cloud; The other phase is called BuildingNet-Mesh, which can also access the mesh data with subgroups. The final PointNeXt-XL model achieves 31.33 PartIoU for BuildingNet-Points segmentation, which marginally beats the MinkNet baseline but significantly improves over the PointNet++ model by over 100\%.

\section{Related Work} 


PointNet\cite{qi2017pointnet} and 3D convolution are two popular building blocks for constructing 3d deep learning models for unstructured point cloud. On one hand, the basic idea of PointNet\cite{qi2017pointnet} is applying MLP layers on each point feature, then, use max pooling to aggregate into global feature. PointNet++ follows by introducing set abstraction to form hierarchical point set features. More recently, PointMLP\cite{ma2022rethinking} proposes Geometric Affine module and Residual Point block as an alternative to extract local point set features.

On the other hand, given the sparse nature of point cloud in 3D space, many researchers have explored 3D sparse convolution to build their 3D models. Submanifold Sparse Convolutional Networks\cite{graham2017submanifold} and Minkowski Convolution Neural Networks\cite{choy20194d} are two noteable models in this direction. Submanifold SparseConv preserves the input/output space structure, while Minkowski SparseConv generalizes it to allow arbitrary output coordinates and sparse kernel.

The BuildingNet paper reported results of PointNet++ and MinkNet on BuildingNet V0 dataset. For the updated v1 dataset, they established a new baseline using MinkRes16UNet34C model. While there is a small difference between V0 and V1 results, it is clear that there is a huge gap between PointNet++ and MinkNet, which affects the corresponding performance on BuildingNet-Mesh when combined with BuildingGNN.

\begin{table}[htb!]
  \centering
    \begin{tabular}{ c | c | c | c }
    \hline
    Method & Dataset & Test PIoU & Test SIoU \\ 
    \hline
    PointNet++ & V0 & 14.1 & 16.7 \\
    \hline
    MinkNet & V0 & 29.9 & 24.3 \\
    \hline
    MinkNet & V1 & 31.2 & 24.1 \\
    \hline
    My PointNeXt-XL & V1 & 31.33 & 22.78 \\
    \hline
    \end{tabular}
\caption{Summary of BuildingNet baseline and my result.}
\label{table:baseline}
\end{table}


With the recent advancement in foundation language models\cite{bommasani2021opportunities}, how to effectively pretrain vision models with self-supervision and leverage multi-modal information has become an important technique to improve supervised vision tasks. CLIP\cite{radford2021learning} and SLIP\cite{mu2022slip} started to learn an aligned representation from text and image pair and has significantly improved zero-shot classification and linear classification. ULIP\cite{xue2022ulip} extends that by generating text, image, and point cloud triplet, and train the point cloud encoder to align its output with text and image encoder outputs. The ULIP author provided pretrained ULIP with PointNeXt-s model on ShapeNet55, and reported promising zero shot classification results on ModelNet40. ULIP-2\cite{xue2023ulip} further extends that by generating text prompts using large multimodal model, and pretrain on larger Objaverse dataset.


PointNeXt\cite{qian2022pointnext} revisited PointNet++ and aim to improve it in three areas: data augmentation, optimization, and model scaling. It also introduces Inverted Residual MLP blocks and receptive field scaling to allow model scaling and reduce receptive field sensitivy issue in PointNet++. PointNeXt, together with ULIP pretraining, has shown really strong performance in various 3D benchmark, such as ScanObjectNN, S3IDS, ShapeNet-Part, and ModelNet40. This encourages more research opportunity on PointNet-based 3D models. 

The main difference between this paper and the original BuildingNet paper is that 

\begin{enumerate}
    \item I use PointNeXt model instead of PointNet++ model, with improved data augmentation, optimization, and model scaling;
    \item I tried various learning strategies, including ULIP pretraining and multi-task learning on both segmentation and classification tasks. More detailed comparison will be discussed in the following sections.
\end{enumerate}

\section{Methods}


Now, we formally define 3D classification and segmentation on point cloud. Given an unordered point set $P = \{p_1, p_2, ..., p_n\}$, where each point $p_i$ is placed in a 3-dimensional space with coordinates $(x_i, y_i, z_i)$, and has a feature $f_i \in \mathbb{R}^d$. A 3D classification model on point cloud predicts a single class label $c$ for $P$. A 3D semantic segmentation model is expected to predict a class label $s_i$ to each point $p_i \in P$. For ULIP pretraining, the model will also receive a list of text prompts, and a list of multi-view images as additional training data. 

\subsection{Review of PointNet++ and PointNeXt}

There are two main types of layer blocks in PointNet++: Set Abstraction and Feature Propagation. PointNeXt extends that with Inverted Residual MLP (InvResMLP). Set Abstraction and InvResMLP are used in the encoder stage.

\textbf{Set Abstraction} aims to reduce number of features by aggregating from groups of neighboring points around the centroid.

\textbf{InvResMLP} has three contributions:

\begin{enumerate}
    \item residual connection, similar as ResNet to reduce vanishing gradient problem;
    \item separable MLP, MLP layers are added after the reduction stage, this is similar to MobileNet to reduce model size;
    \item  the output channel of second MLP layer can be expanded for model scaling.
\end{enumerate}

\textbf{Feature Propagation} is used in the segmentation decoder stage to scale the encoder feature back to original point set size.

For classification, fully connected layers are used to propagate encoder feature to class logits.

\begin{figure}[htb!]
\centering
\includegraphics[width=0.8\linewidth]
               {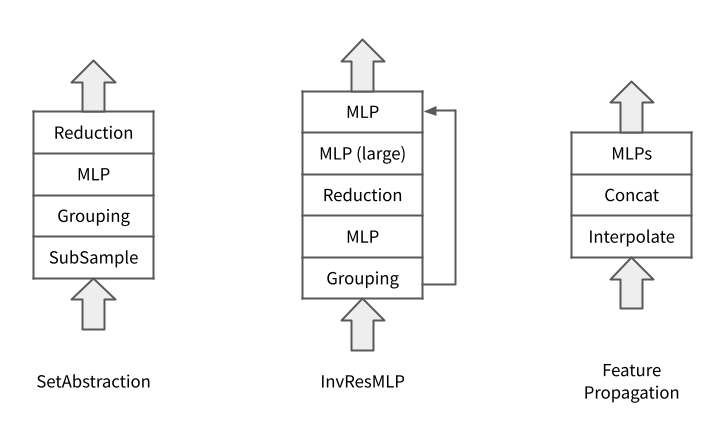}
\caption{Model Layer Blocks in PointNet++/PointNeXt.}
\label{fig:model_layer_blocks}
\end{figure}

These three layer blocks can be futher broken down into following layers:

\textbf{Sampling Layer.} Given input points $\{p_1, p_2, ..., p_n\}$, the sampling layer uses farthest point sampling (FPS) to choose a subset of points $\{p_{i_1}, p_{i_2}, ..., p_{i_m}\}$, such that $p_{i_j}$ is the most distant point from $\{p_{i_1}, p_{i_2}, ..., p_{i_{j-1}}\}$. The sampled point set size is controlled by stride parameter $s$, i.e., $m = n // s$. FPS has better coverage than uniform sampling. The subset of points become the centroid points to the next grouping layer.

\textbf{Grouping Layer.} Taking the original input points and centroid points from sampling Layer, the grouping layer groups neighbors of centroids. Ball query is typically used to find neighbors within a radius to the centroid point. The radius $r$ determines the receptive field and is sensitive to the density of point cloud, especially when the point cloud is subsampled. In PointNet++, the initial value $r$ will double after each sampling layer. As mentioned in PointNeXt paper, the radius value is very dataset dependent, especially if the dataset is voxelized with different voxel size. We maintain the same ratio between voxel size and radius value as PointNeXt to achieve reasonable performance.

MLP Layer operates locally on each centroid, taking relative coordinate $p_j - p_i$ as input. Reduction Layer aggregates features from neighbors (e.g. max-pooling). Finally, Feature Propagation Layer uses inverse distance weighted average based on $k$ nearest neighbors to interpolate features of sampled points back to original points.

Our baseline models will be training PointNeXt from scratch. I use cross entropy loss for classification, and pixel-wise weighted cross entropy loss for segmentation. The weight is calculated by inverse logarithm of label frequency from each task.

From there, I explore how different learning strategies can help on various tasks in three categories: 1. transfer learning; 2. ULIP pretraining; 3. multi-task training.

\subsection{Transfer learning from a different dataset}

In 2D image modeling, transfer learning from a different dataset is commonly used to boost the training performance and reduce the dataset size for finetuning \cite{huh2016makes}.Given ULIP has outstanding performance in ModelNet40 benchmark\footnote{\url{https://paperswithcode.com/sota/3d-point-cloud-classification-on-modelnet40}}, we want to first try to load pretrained ULIP+PointNeXt checkpoint from ShapeNet for 3d classification task. Note that there are many shared classes between ShapeNet55 and ModelNet40, but there isn't any class related to buildings. I load pretrained ULIP with PointNeXt-s backbone from ShapeNet55\cite{chang2015shapenet} and finetune it on 3d classification task to see whether transfer learning helps in such setting.

For 3d segmentation task, the PointNeXt model has achieved top performance on S3DIS, a 3d indoor scene datasets. Again, there is a huge difference between S3DIS (indoor scene) and BuildingNet (building exteriors). I load the same backbone for the PointNext-s segmentation model and finetune it on the BuildingNet dataset.

\subsection{Using ULIP as a pretraining framework}

Remind that the ULIP framework freezes the text and image encoders, and align point cloud encoder feature with text and image encoder features with a cross-modal contrastive loss. At test time, it computes the point cloud encoder feature and finds the closest text encoder feature among all categories to predict the class label.

For finetuning, I load the best ULIP checkpoint and finetune it in the corresponding model. The PointNeXt-s encoders for classification and segmentation uses difference block sizes and stride sizes. For baseline, I train the classification model from scratch with both classification backbone and segmentation backbone for comparison, but here I only pretrain ULIP with segmentation backbone for consistency.

\begin{figure}[htb!]
\centering
\includegraphics[width=0.8\linewidth]
               {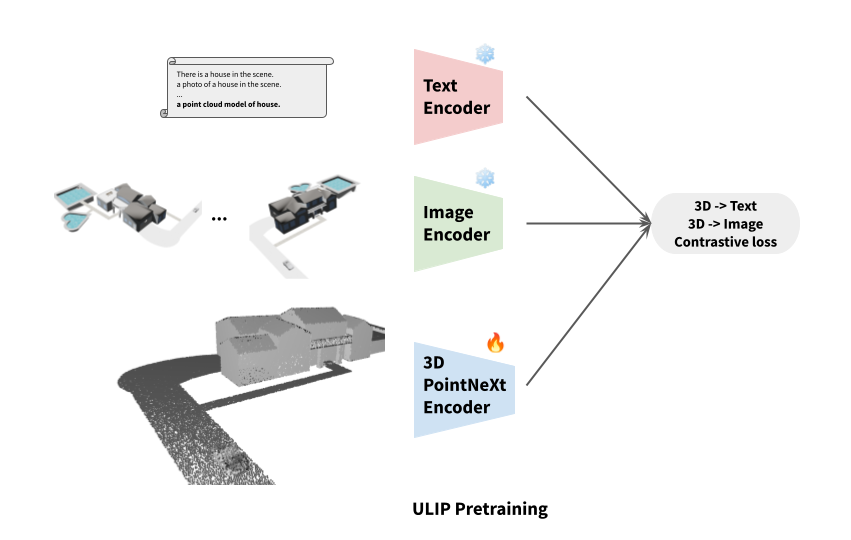}
\caption{ULIP pretraining process on BuildingNet.}
\label{fig:ulip_pretraining}
\end{figure}

\subsection{Multi-task learning}

Lastly, I train a dual-task model that does both classification and segmentation with a shared backbone, repeat the experiment of training from scratch and loading pretrained checkpoint from ULIP.

I use weighted sum to balance between tasks:

\begin{equation}
    L_{\text{total}} = \beta L_{\text{classification}} + (1 - \beta) L_{\text{segmentation}}
\end{equation}

\section{Dataset and Features}

Each 3D building model comes with the following data:

\begin{itemize}
    \item Name: The name is prefixed with building class and subclass, e.g., in "COMMERCIALcastle\_mesh0365", "COMMERCIAL" is the building class, "castle" is the subclass. I use the subclass ("castle") as the building type classification label.
    \item Feature: Coordinates $(x, y, z)$, Normal $(nx, ny, nz)$, Color $(r, g, b)$. All coordinates are prenormalized to $[-0.5, 0.5]$ and downsampled to 100,000 points, all features are prenormalized to $[0, 1]$. Following PointNeXt, I use $y - min(y)$ to generate the heights feature to help model differentiate between roof and ground.
    \item Per-point class index for segmentation. For challenge submission, labels are only provided for training and validation splits.
    \item 3DWarehouse IDs and links to original models.
\end{itemize}

\subsection{Text-image-point triplet generation}

ULIP pretraining requires text prompts and multi-view images to be generated with the point clouds. I generated 64 text prompts for each 3d building, 63 text prompts were generated using prompt template from \cite{gu2021open}, then I added a dedicated prompt "a point cloud model of {category}". Each text prompt will be sent to a pretrained SLIP text encoder, then I use average pooling to get the final text encoder feature. For multi-view images, I use the same renderer script as ULIP \footnote{\url{https://github.com/panmari/stanford-shapenet-renderer}}, but only generated 8 RGB images and depth maps per every 45 degrees to save storage space. This generates 16 images per building in total. During pretraining, I randomly selected one image and send that to a pretrained SLIP image ViT encoder to generate the image encoder feature.

\subsection{Point cloud preprocessing}

I perform all 3d point cloud preprocessing on the fly in the dataset loader to maximize the data diversity for model training. It can be divided into following stages: pre-voxelize data augmentation, voxelization, and post-voxelize data augmentation.

\textbf{Pre-voxelize data augmentation.} I follow BuildingNet baseline to apply random rotation before voxelization for data augmentation. Random rotation is mentioned as a strong data augmentation in PointNeXt as well, however, they didn't apply it to PointNeXt-s as they see performance drop. BuildingNet only has ~2k 3d models, so adding this is critical to avoid overfitting. The entire dataset will also be looped 12 times for each epoch.

\textbf{Voxelization.} It is a common technique to group 3D point clouds into voxel grids. The voxel grids can be further downsampled to fit into the model. At training time, a random point from each voxel grid will be selected for model inputs. At test time, an exhaustive sampling will generate a list of sub-clouds that each sub-cloud contains one point from each voxel, so that the model can run inference on the entire cloud. The voxelized point cloud will be further downsampled to a fixed size.

\begin{figure}[htb!]
\centering
\includegraphics[width=0.8\linewidth]
               {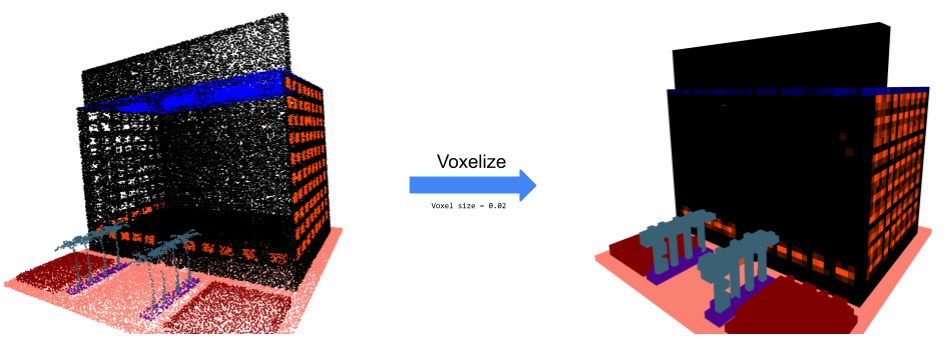}
\caption{Voxelize Visualization.}
\label{fig:voxelize}
\end{figure}

\textbf{Post-voxelize data augmentation.} I adopt all data augmentation used by PointNeXt model in post-voxelize data augmentation, except for random rotation. This includes color auto contrast, random scaling, jittering, and color drop. Color auto contrast automatically adjust color contrast \cite{zhao2021point}, random scaling randomly scale the entire point cloud by a small factor, jittering randomly add independent noise to each point, and color drop randomly drop the rgb color feature to force the model to learn more from other features and the geometry relationship between points.

\subsection{Data statistical analysis}

BuildingNet is a challenging dataset. The reported metrics from MinkNet baseline is significantly lower than other 3D semantic segmentation datasets, such as S3DIS \footnote{\url{https://paperswithcode.com/sota/semantic-segmentation-on-s3dis}} and Waymo Open Dataset \footnote{\url{https://waymo.com/open/challenges/2022/3d-semantic-segmentation}}. I perform data statistical analysis to understand why this is so challenging before diving into experiments.

\begin{figure}[htb!]
\centering
\includegraphics[width=0.8\linewidth]
               {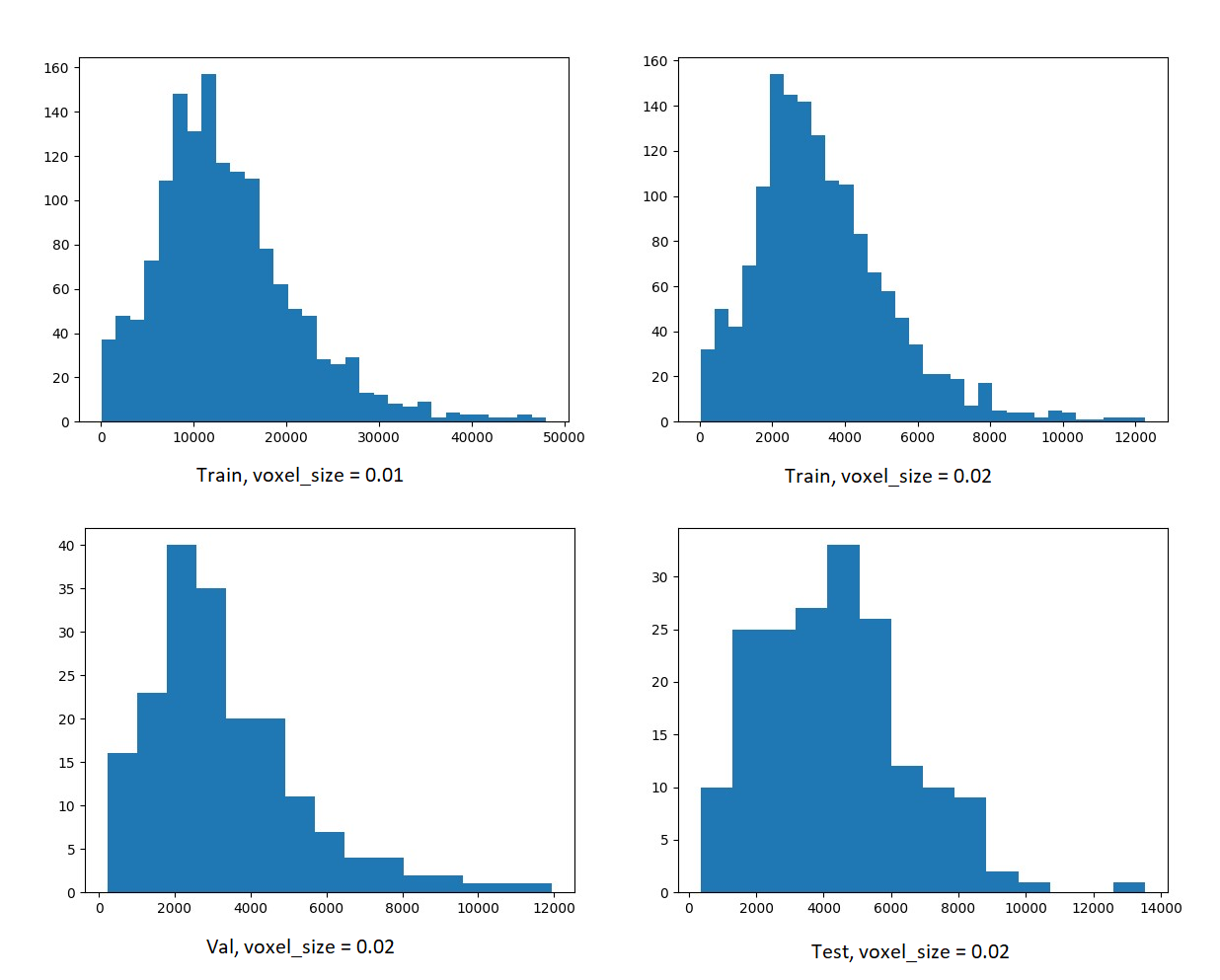}
\caption{Histogram of number of voxels per point cloud model.}
\label{fig:voxel_histogram}
\end{figure}

Figure \ref{fig:voxel_histogram} shows the distribution of number of voxels on different voxel sizes and data splits. We can see that as the voxel size becomes smaller, the number of voxels grows dramatically. More importantly, I notice a distribution difference between train/val splits and test split, Both train and val splits have peaks around 2000 voxels, but the test split peaks at 5000 voxels. I empirically choose voxel size and sample size to be 0.02 and 12,500 for training experiments, and set ball query radius to 0.05.

\begin{figure}[htb!]
\centering
\includegraphics[width=0.8\linewidth]
               {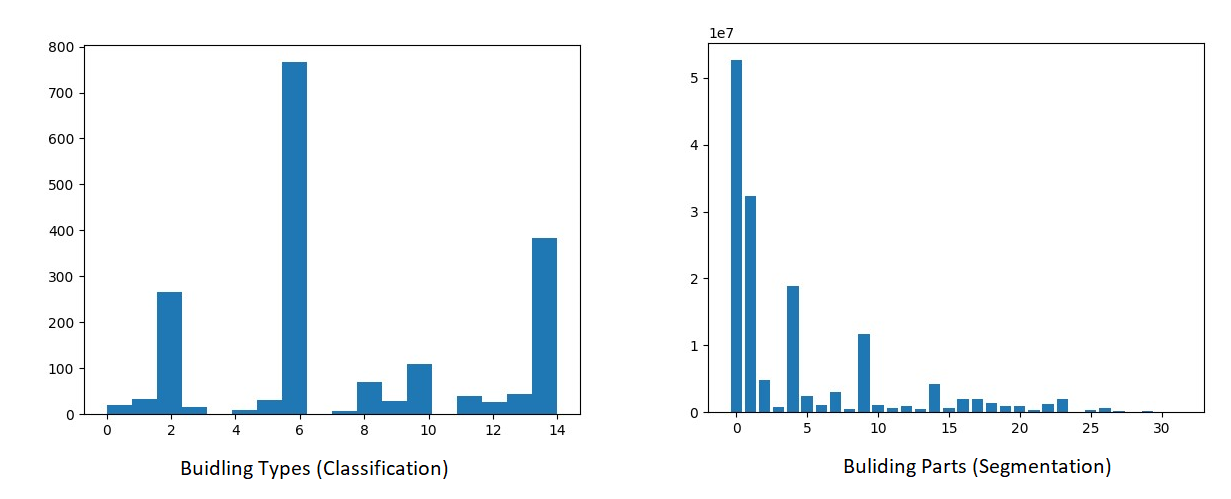}
\caption{Histogram of number of labels for building type classification and building segmentation segmentation.}
\label{fig:label_histogram}
\end{figure}

BuildingNet is a small dataset. There are 1480 buildings in train split, 187 buildings in val split, and 181 buildings in test split. While it has 15 building type classes and 31 building part classes. In Figure \ref{fig:label_histogram}, we can see that the class distributions are extremely unbalanced for both classification and segmentation. In segmentation labels, there is a massive
number of points that come with segmentation class index 0, i.e., ”unspecified”, which means those points are not labeled. We have excluded class 0 from data, loss function, and evaluation metrics to avoid any confusion to the model.

\section{Experiments}

\subsection{Experiment Plan}

Table \ref{table:learning_strategies} shows all combinations of learning strategies we have discussed. I use PointNeXt-s model for all experiments, except for the final model scale-up. Training epoch is fixed at 100, learning rate is 0.01, cosine decay is used for learning rate schedule for all experiments without specially mentioned. For multi-task training, we experimented with $\beta = 0.01$ and $\beta = 0.03$.

\begin{table}[htb!]
  \centering
    \begin{tabular}{ c | c | c }
    \hline
    Id & Pretrain & Task \\ 
    \hline
    1 & Scratch & Classification \\
    \hline
    1' & Scratch (Seg backbone) & Classification \\
    \hline
    2 & Scratch & Segmentation \\
    \hline
    3 & Scratch & ULIP \\
    \hline
    4 & ULIP(ShapeNet) & Classification \\
    \hline
    5 & PointNeXt(S3DIS) & Segmentation \\
    \hline
    6 & ULIP(BuildingNet) & Classification \\
    \hline
    7 & ULIP(BuildingNet) & Segmentation \\
    \hline
    8 & Scratch & Class+Seg \\
    \hline
    9 & ULIP(BuildingNet) & Class+Seg \\
    \hline
    \end{tabular}
\caption{Summary of learning strategies.}
\label{table:learning_strategies}
\end{table}

\subsection{Evaluation Metrics}

For 3d building type classification, we simply calculate overall accuracy for evaluation.

\begin{equation}
    \text{accuracy} = \frac{\text{\# true positive buildings}}{\text{\# all buildings}}
\end{equation}

For 3d building part segmentation, we mainly use PartIoU metrics\cite{mo2019partnet}, which calculates the average of intersection over union (IoU) for each semantic part across all buildings. The BulidingNet challenge also calculates ShapeIoU, which calculates the average of IoU per building.

\begin{equation}
  \text{PartIoU} = \frac{1}{N_p}(\sum_{B_i \in B}\text{IoU}_{p_i, B_i})
\end{equation}
where $p$ is a building part, e.g. roof, wall, $N_p$ is the number of part classes, $\text{IoU}_{p_i, B_i}$ calculates the IoU for part $p_i$ in buliding $B_i$.

Last but not least, for dual-task training, we use the harmonic mean of classification accuracy and segmentation PartIoU to select the best checkpoint.

\begin{equation}
    \text{Harmonic mean} = \frac{2}{\text{accuracy}^{-1} + \text{PartIoU}^{-1}}
\end{equation}

Next, we will discuss experiment results by tasks.

\subsection{3D Building Type Classification}

\begin{figure}[htb!]
\centering
\includegraphics[width=0.8\linewidth]
               {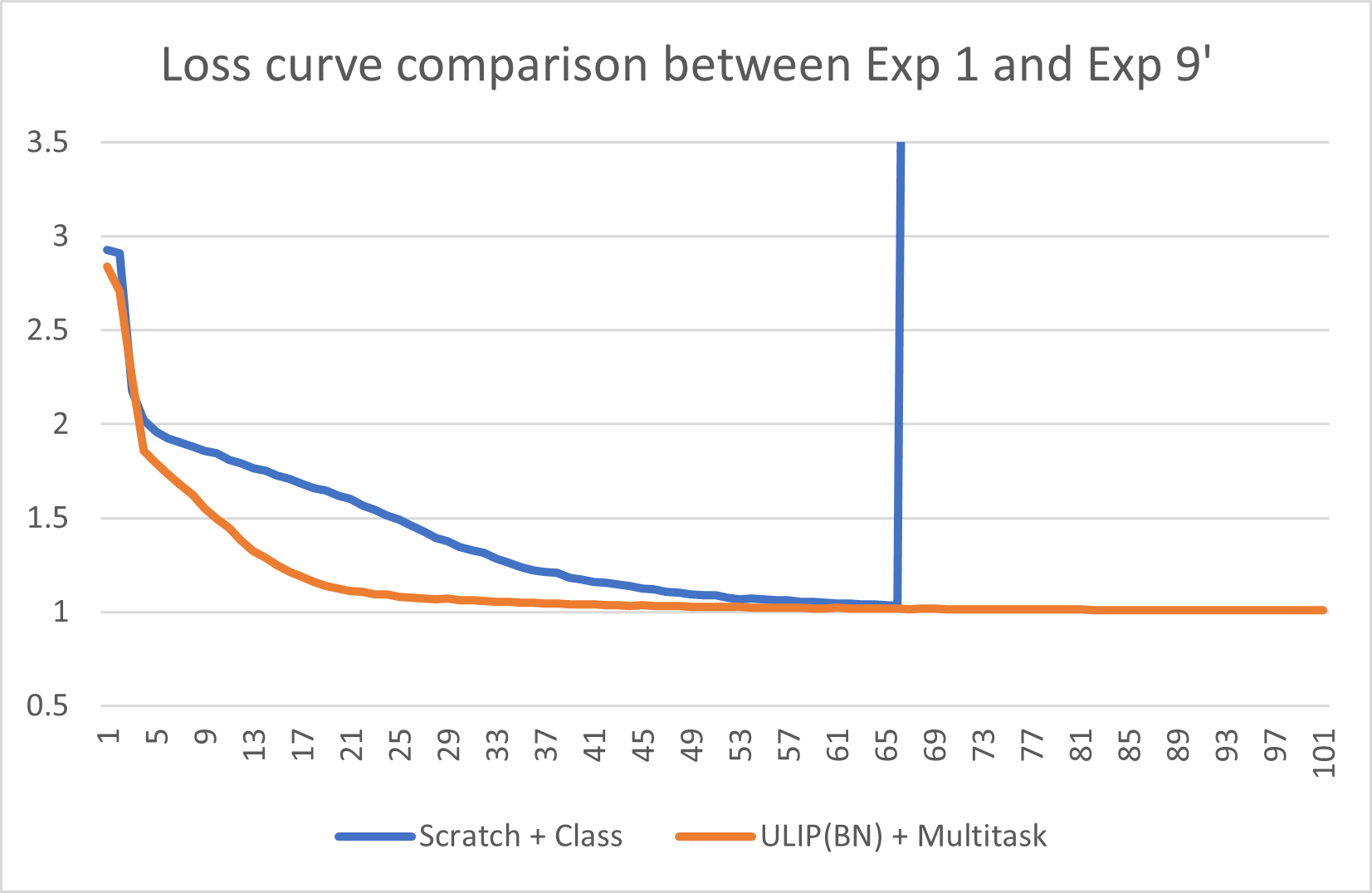}
\caption{Classification loss comparison between Exp1 and Exp9'.}
\label{fig:classification_loss}
\end{figure}

Table \ref{table:classification_results} shows the experiment results for classification. The training accuracy is recorded at the best validation checkpoint. Training from scratch using classification backbone (Exp 1), and Multi-task training with ULIP pretraining (Exp 9') achieves top two performance. Figure \ref{fig:classification_loss} compares the classification loss curve between two models. The classification loss for training from scratch got exploded to NaN at epoch 66, and the best checkpoint was found early in epoch 16. The classification loss for multi-task training with ULIP pretraining went down more efficiently.

For ULIP pretraining, training accuracy was not reported. Training ULIP from scratch (Exp 3) is lower than other experiments, because it only uses a projection matrix to project encoder features into a hidden space for feature alignment. However, further finetuning it (Exp 6) doesn't bringing the performance better than training from scratch (Exp 1').

The multi-task training reacts differently to the task weight $\beta$ when training from scratch and loading ULIP pretrained checkpoint, this suggests that when loading ULIP pretrained checkpoint, the model already learns some information about classification and requires less finetuning on it.

\begin{figure}[htb!]
\centering
\includegraphics[width=0.6\linewidth]
               {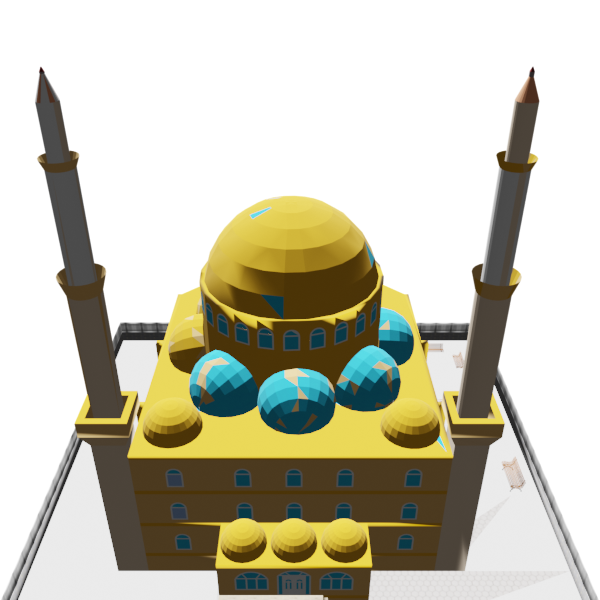}
\caption{A rendered example of mosque.}
\label{fig:mosque}
\end{figure}

I further look into the per-class accuracy for the multi-task model (Exp 9'). The validation set is so small that most classes only have 1-2 buildings. The model is only able to produce metrics on the 6 dominant classes of the BuildingNet dataset, each has more than 40 buildings. The model archieves 79.49 accuracy on house but 50 on villa, which indicates the model got confused villa as many examples look similar to house. Surprisingly, the model achieves a high accuracy score on mosque, probably because of its unique tower shape (Figure \ref{fig:mosque}). It is clear that the model needs more training and validation data to achieve reasonable building type classification performance.

\begin{table}[htb!]
  \centering
    \begin{tabular}{ c | c | c | c }
    \hline
    Id & Model & Train Acc & Val Acc \\ 
    \hline
    1 & Scratch+Class & 61.97 & \textbf{60.43} \\
    \hline
    1' & Scratch+Seg & 58.83 & 58.82 \\
    \hline
    3 & Scratch+ULIP & N/A & 54.55 \\
    \hline
    4 & ULIP(ShapeNet)+Class & 68.23 & 56.68 \\
    \hline
    6 & ULIP(BN)+Seg & 54.92 & 57.22 \\
    \hline
    8 & Scratch+Multitask & \\
    & $\beta = 0.03$ & 99.84 & 58.29 \\
    \hline
    8' & Scratch+Multitask & \\
    & $\beta = 0.01$ & 99.4 & 56.68\\
    \hline
    9 & ULIP(BN)+Multitask & \\
    & $\beta = 0.03$ & 99.79 & 57.75 \\
    \hline
    9' & ULIP(BN)+Multitask & \\
    & $\beta = 0.01$ & 97.93 & \textbf{59.36} \\
    \hline
    \end{tabular}
\caption{Summary of classification evaluation results.}
\label{table:classification_results}
\end{table}

\begin{table}[htb!]
  \centering
    \begin{tabular}{ c | c | c }
    \hline
    Building Type & Accuracy & \# buildings \\
    \hline
    house & 79.49 & 766 \\
    \hline
    villa & 50. & 384 \\
    \hline
    church & 63.16 & 265 \\
    \hline
    office building & 50 & 109 \\
    \hline
    mosque & 72.73 & 70 \\
    \hline
    temple & 28.57 & 45 \\
    \hline
    \end{tabular}
\caption{Summary of accuracy per class.}
\label{table:accuracy_per_class}
\end{table}

\subsection{3D Building Part Segmentation}

Table \ref{table:segmentation_results} shows the experiment results for 3d building part segmentation. Both segmentation model and multitask model with ULIP pretraining achieves the best performance. Using ULIP as a pretraining framework is helpful for the segmentation task. Transfer learning from S3DIS doesn't really help because of the domain difference between indoor scene and building exteriors.


\begin{table}[htb!]
  \centering
    \begin{tabular}{ c | c | c | c }
    \hline
    Id & Model & Train PIoU & Val PIoU \\ 
    \hline
    2 & Scratch+Seg & 55.44 & 31.66 \\
    \hline
    5 & PN(S3DIS)+Seg & 34.65 &  26.36 \\
    \hline
    7 & ULIP(BN)+Seg & 56.1 & \textbf{32.09} \\
    \hline
    8 & Scratch+Multitask & \\
    & $\beta = 0.03$ & 51.52 & 29.77 \\
    \hline
    8' & Scratch+Multitask & \\
    & $\beta = 0.01$ & 52.14 & 29.78 \\
    \hline
    9 & ULIP(BN)+Multitask & \\
    & $\beta = 0.03$ & 45.81 & 29.72 \\
    \hline
    9' & ULIP(BN)+Multitask & \\
    & $\beta = 0.01$ & 50 & \textbf{31.68} \\
    \hline
    \end{tabular}
\caption{Summary of segmentation evaluation results.}
\label{table:segmentation_results}
\end{table}

The ULIP framework is agnostic to point cloud backbone architecture. I also train ULIP+MinkNet by finetuning from segmentation baseline checkpoint for only 50 epochs. The result in Table \ref{table:ulip_results} shows that transfer learning from segmentation to classification task may also be useful.

\begin{table}[htb!]
  \centering
    \begin{tabular}{ c | c | c }
    \hline
     & MinkNet & PointNeXt-s \\
    \hline
    Val Top-1 accuracy & 55.61 & 54.54 \\
    \hline
    Val Top-5 accuracy & 85.56 & 85.56 \\
    \hline
    \end{tabular}
\caption{ULIP+MinkNet v.s. ULIP+PointNeXt-s.}
\label{table:ulip_results}
\end{table}

\subsection{Final model scale-up}

\begin{figure}[htb!]
\centering
\includegraphics[width=0.8\linewidth]
               {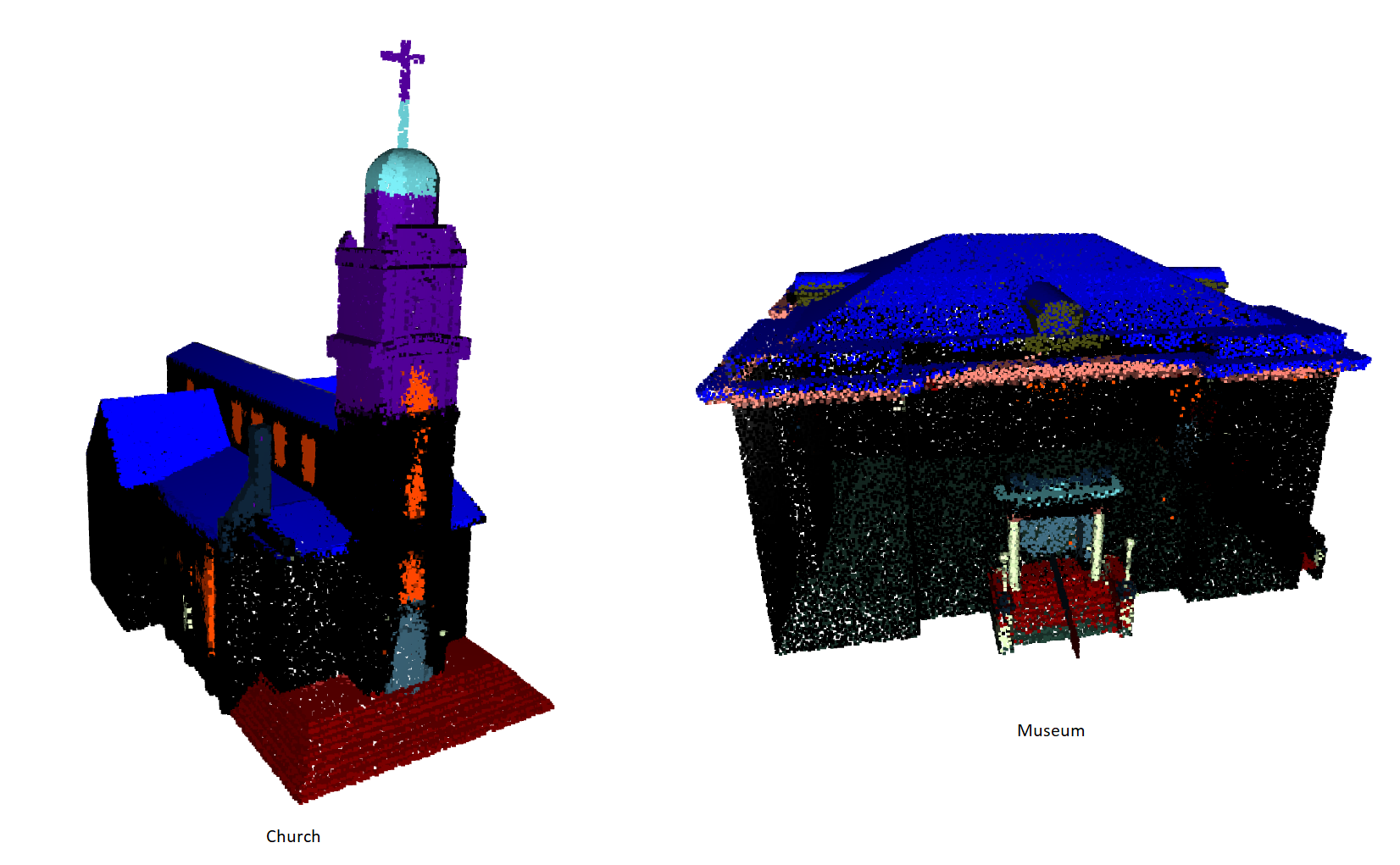}
\caption{Visualization of PointNeXt-XL segmenation results on test split.}
\label{fig:seg_visualization}
\end{figure}

The final model is to scale up model to PointNeXt-XL and train a model for the BuildingNet challenge submission. I updated voxel size and sample size to be 0.01 and 40,000, and set ball query radius to 0.025. Due to the time and resource constraint, I only train one PointNeXt-XL model from scratch on 8xA100 multi-gpu in order to submit the test result in time. Table \ref{table:minknet-pointnext-xl} compares between MinkNet and PointNext-XL. Note that both optimizers use momentum and weight decay. Looking at the segmentation visualization of PointNeXt-XL test result (Figure \ref{fig:seg_visualization}), while the prediction result is still noisy, the model can already learn most building parts, such as window, wall, roof, etc.

Table \ref{table:per_class_iou} compares per-class segmentation IoU between PointNeXt-s (ULIP(BN)+Seg), PointNeXt-XL (Scratch+Seg), and MinkNet\footnote{This came from a finetune training from MinkNet baseline checkpoint.}. PointNeXt-s wins 4 classes, PointNeXt-XL wins 11 classes, MinkNet wins 15 classes.

\begin{table}[htb!]
  \centering
    \begin{tabular}{ c | c | c }
    \hline
    Model Type & MinkNet & PointNeXt-XL \\
    \hline
    Pre-voxelize & \multicolumn{2}{c}{Random rotation} \\
    \hline
    Voxel Size & \multicolumn{2}{c}{0.01} \\
    \hline
    Sample Size & 100,000 & 40,000 \\
    \hline
    Post-voxelize & None & color auto contrast,  \\
    Data & & random scaling, \\
    Augmentation & & jittering, color drop \\
    \hline
    Loop Size & \multicolumn{2}{c}{12} \\
    \hline
    Training Epoch & 200 & 100 \\
    \hline
    Loss & \multicolumn{2}{c}{weighted cross entropy} \\
    \hline
    LR & \multicolumn{2}{c}{0.01} \\
    \hline
    Scheduler & \multicolumn{2}{c}{cosine} \\
    \hline
    Optimizer & SGD & Adam \\
    \hline
    \end{tabular}
\caption{Comparison between Minknet and PointNeXt-XL}
\label{table:minknet-pointnext-xl}
\end{table}

\begin{table}[htb!]
  \centering
    \begin{tabular}{ c | c | c | c }
    \hline
    & PointNeXt-s & PointNeXt-XL & MinkNet \\
    \hline
    PartIoU &&& \\
    (train) & 56.1 & \textbf{74.25} & 65.72 \\
    \hline
    PartIoU &&& \\
    (val) & 32.09 & \textbf{34.68} & 34 \\
    \hline
    PartIoU &&&  \\
    (test) & 29.56 & \textbf{31.33} & 31.2 \\
    \hline
    ShapeIoU &&&  \\
    (test) & 21.76 & 22.78 & \textbf{24.1} \\
    \hline
    \end{tabular}
\caption{PointNeXt-s v.s. PointNeXt-XL v.s. MinkNet}
\label{table:per_class_iou}
\end{table}

\begin{table}[htb!]
  \centering
    \begin{tabular}{ c | c | c | c}
    \hline
    class & PointNeXt-s & PointNeXt-XL & MinkNet  \\ 
    \hline
    wall & 59.21 & 64.55 & \textbf{65.31} \\
    \hline
    window & 30.45 & \textbf{32.63} & 29.89 \\
    \hline
    vehicle & 54.05 & \textbf{55.29} & 45.27 \\
    \hline
    roof & 72.9 & \textbf{75.79} & 72.16 \\
    \hline
    plant & 71.86 & 75.12 & \textbf{77.14} \\
    \hline
    door & 9.79 & 17.59 & \textbf{18.66} \\
    \hline
    tower & 35.66 & \textbf{50.49} & 42.88 \\
    \hline
    furniture & 25.96 & \textbf{37.84} & 33.87 \\
    \hline
    ground & 67.95 & 65.79 & \textbf{74.27} \\
    \hline
    beam & 36.82 & \textbf{42.44} & 29.46 \\
    \hline
    stairs & 19.52 & \textbf{32.83} & 29.18 \\
    \hline
    column & 25.7 & 25.52 & \textbf{34.80} \\
    \hline
    banister & 19.58 & 26.46 & \textbf{35.66} \\
    \hline
    floor & 40.22 & 40.29 & \textbf{41.70}\\
    \hline
    chimney & \textbf{38.03} & 36.17 & 33.75 \\
    \hline
    ceiling & \textbf{26.51} & 20.43 & 22.97 \\
    \hline
    fence & 50.32 & 39.81 & \textbf{54.65} \\
    \hline
    pool & 53.85 & 51.21 & \textbf{67.86} \\
    \hline
    corridor & \textbf{16.64} & 15.56 & 14.32\\
    \hline
    balcony & 21.91 & 22.19 & \textbf{28.47} \\
    \hline
    garage & 10.51 & 23.9 & \textbf{30.65} \\
    \hline
    dome & 56.22 & \textbf{61.64} & 52.51 \\
    \hline
    road & 44.84 & \textbf{51.43} & 50.23 \\
    \hline
    gate & 0 & \textbf{6.82} & 0 \\
    \hline
    parapet & 2.23 & 1.74 & \textbf{2.81} \\
    \hline
    buttress & 18.96 & 24.59 & \textbf{28.90} \\
    \hline
    dormer & 2.19 & 6.87 & \textbf{10.72} \\
    \hline
    lighting & 0 & 0.17 & \textbf{16.72} \\
    \hline
    arch & 1.19 & \textbf{4.38} & 1.36 \\
    \hline
    awning & \textbf{0.19} & 0.1 & 00.15 \\
    \hline
    shutters & 0 & 0 & 0 \\
    \hline
    \end{tabular}
\caption{Per-class IoU comparison}
\label{table:per_class_iou}
\end{table}

\pagebreak

\section{Conclusion}

PointNet and SparseConv are two building blocks in the 3d deep learning toolkit. The comparison between PointNeXt and MinkNet show that both can achieve reasonable performance on a challenging dataset like BuildingNet. Multi-task training with multi-modal pretraining consistently achieves top performance on building type classification and building part segmentation. Scaling up PointNeXt model achieves the best model performance in segmentation task. What's more, we can reduce ULIP training time by loading a good segmentation model. Not just labeling on 3D is expansive, but also training on 3D requires significant accelerator resource. There is a lot of head room to continue research in 3D deep learning and find more efficient ways to boost data quality and model performance.

\vspace{100pt}

\pagebreak

\section{Acknowledgements}
\label{sec:Acknowledgements}

This Stanford CS231n project is mentored by Alberto Tono, who helped review proposal, milestone, and final report, and provided suggestions during project development. I want to thank Chen Xia, and Juan Miguel Navarro Carranza for discussions and contributions before project milestone. All the code, experiments, and the final report are authored and completed by the author. The codebase exists of 4k+ lines of custom code while reusing PointNeXt\footnote{\url{https://github.com/guochengqian/PointNeXt}}, ULIP\footnote{\url{https://github.com/salesforce/ULIP}}, and buildingnet\_dataset\footnote{\url{https://github.com/buildingnet/buildingnet_dataset}} repos. 

\section{Appendix}
\subsection{Complete list of building subtypes}

\begin{itemize}[noitemsep,topsep=0pt,parsep=0pt,partopsep=0pt]
\item castle
\item cathedral
\item church
\item city hall
\item factory
\item hotel building
\item house
\item monastery
\item mosque
\item museum
\item office building
\item palace
\item school building
\item temple
\item villa
\end{itemize}

{\small
\bibliographystyle{ieee_fullname}
\bibliography{main}

\begin{thebibliography}{10}\itemsep=-1pt

\bibitem{armeni_cvpr16}
Iro Armeni, Ozan Sener, Amir~R. Zamir, Helen Jiang, Ioannis Brilakis, Martin
  Fischer, and Silvio Savarese.
\newblock 3d semantic parsing of large-scale indoor spaces.
\newblock In {\em Proceedings of the IEEE International Conference on Computer
  Vision and Pattern Recognition}, 2016.

\bibitem{bommasani2021opportunities}
Rishi Bommasani, Drew~A Hudson, Ehsan Adeli, Russ Altman, Simran Arora, Sydney
  von Arx, Michael~S Bernstein, Jeannette Bohg, Antoine Bosselut, Emma
  Brunskill, et~al.
\newblock On the opportunities and risks of foundation models.
\newblock {\em arXiv preprint arXiv:2108.07258}, 2021.

\bibitem{chang2015shapenet}
Angel~X Chang, Thomas Funkhouser, Leonidas Guibas, Pat Hanrahan, Qixing Huang,
  Zimo Li, Silvio Savarese, Manolis Savva, Shuran Song, Hao Su, et~al.
\newblock Shapenet: An information-rich 3d model repository.
\newblock {\em arXiv preprint arXiv:1512.03012}, 2015.

\bibitem{choy20194d}
Christopher Choy, JunYoung Gwak, and Silvio Savarese.
\newblock 4d spatio-temporal convnets: Minkowski convolutional neural networks.
\newblock In {\em Proceedings of the IEEE/CVF conference on computer vision and
  pattern recognition}, pages 3075--3084, 2019.

\bibitem{fedorova2021synthetic}
Stanislava Fedorova, Alberto Tono, Meher~Shashwat Nigam, Jiayao Zhang,
  Amirhossein Ahmadnia, Cecilia Bolognesi, and Dominik~L Michels.
\newblock Synthetic 3d data generation pipeline for geometric deep learning in
  architecture.
\newblock {\em arXiv preprint arXiv:2104.12564}, 2021.

\bibitem{graham2017submanifold}
Benjamin Graham and Laurens Van~der Maaten.
\newblock Submanifold sparse convolutional networks.
\newblock {\em arXiv preprint arXiv:1706.01307}, 2017.

\bibitem{gu2021open}
Xiuye Gu, Tsung-Yi Lin, Weicheng Kuo, and Yin Cui.
\newblock Open-vocabulary object detection via vision and language knowledge
  distillation.
\newblock {\em arXiv preprint arXiv:2104.13921}, 2021.

\bibitem{handa2015scenenet}
A Handa, V Patraucean, V Badrinarayanan, S Stent, and R Cipolla.
\newblock Scenenet: understanding real world indoor scenes with synthetic data.
  arxiv preprint (2015).
\newblock {\em arXiv preprint arXiv:1511.07041}, 3, 2015.

\bibitem{huh2016makes}
Minyoung Huh, Pulkit Agrawal, and Alexei~A Efros.
\newblock What makes imagenet good for transfer learning?
\newblock {\em arXiv preprint arXiv:1608.08614}, 2016.

\bibitem{liao2022kitti}
Yiyi Liao, Jun Xie, and Andreas Geiger.
\newblock Kitti-360: A novel dataset and benchmarks for urban scene
  understanding in 2d and 3d.
\newblock {\em IEEE Transactions on Pattern Analysis and Machine Intelligence},
  2022.

\bibitem{ma2022rethinking}
Xu Ma, Can Qin, Haoxuan You, Haoxi Ran, and Yun Fu.
\newblock Rethinking network design and local geometry in point cloud: A simple
  residual mlp framework.
\newblock {\em arXiv preprint arXiv:2202.07123}, 2022.

\bibitem{mo2019partnet}
Kaichun Mo, Shilin Zhu, Angel~X Chang, Li Yi, Subarna Tripathi, Leonidas~J
  Guibas, and Hao Su.
\newblock Partnet: A large-scale benchmark for fine-grained and hierarchical
  part-level 3d object understanding.
\newblock In {\em Proceedings of the IEEE/CVF conference on computer vision and
  pattern recognition}, pages 909--918, 2019.

\bibitem{mu2022slip}
Norman Mu, Alexander Kirillov, David Wagner, and Saining Xie.
\newblock Slip: Self-supervision meets language-image pre-training.
\newblock In {\em Computer Vision--ECCV 2022: 17th European Conference, Tel
  Aviv, Israel, October 23--27, 2022, Proceedings, Part XXVI}, pages 529--544.
  Springer, 2022.

\bibitem{qi2017pointnet}
Charles~R Qi, Hao Su, Kaichun Mo, and Leonidas~J Guibas.
\newblock Pointnet: Deep learning on point sets for 3d classification and
  segmentation.
\newblock In {\em Proceedings of the IEEE conference on computer vision and
  pattern recognition}, pages 652--660, 2017.

\bibitem{qian2022pointnext}
Guocheng Qian, Yuchen Li, Houwen Peng, Jinjie Mai, Hasan Hammoud, Mohamed
  Elhoseiny, and Bernard Ghanem.
\newblock Pointnext: Revisiting pointnet++ with improved training and scaling
  strategies.
\newblock {\em Advances in Neural Information Processing Systems},
  35:23192--23204, 2022.

\bibitem{radford2021learning}
Alec Radford, Jong~Wook Kim, Chris Hallacy, Aditya Ramesh, Gabriel Goh,
  Sandhini Agarwal, Girish Sastry, Amanda Askell, Pamela Mishkin, Jack Clark,
  et~al.
\newblock Learning transferable visual models from natural language
  supervision.
\newblock In {\em International conference on machine learning}, pages
  8748--8763. PMLR, 2021.

\bibitem{selvaraju2021buildingnet}
Pratheba Selvaraju, Mohamed Nabail, Marios Loizou, Maria Maslioukova, Melinos
  Averkiou, Andreas Andreou, Siddhartha Chaudhuri, and Evangelos Kalogerakis.
\newblock Buildingnet: Learning to label 3d buildings.
\newblock In {\em Proceedings of the IEEE/CVF International Conference on
  Computer Vision}, pages 10397--10407, 2021.

\bibitem{sun2020scalability}
Pei Sun, Henrik Kretzschmar, Xerxes Dotiwalla, Aurelien Chouard, Vijaysai
  Patnaik, Paul Tsui, James Guo, Yin Zhou, Yuning Chai, Benjamin Caine, et~al.
\newblock Scalability in perception for autonomous driving: Waymo open dataset.
\newblock In {\em Proceedings of the IEEE/CVF conference on computer vision and
  pattern recognition}, pages 2446--2454, 2020.

\bibitem{tono2022vitruvio}
Alberto Tono and Martin Fischer.
\newblock Vitruvio: 3d building meshes via single perspective sketches.
\newblock {\em arXiv preprint arXiv:2210.13634}, 2022.

\bibitem{xue2022ulip}
Le Xue, Mingfei Gao, Chen Xing, Roberto Mart{\'\i}n-Mart{\'\i}n, Jiajun Wu,
  Caiming Xiong, Ran Xu, Juan~Carlos Niebles, and Silvio Savarese.
\newblock Ulip: Learning unified representation of language, image and point
  cloud for 3d understanding.
\newblock {\em arXiv preprint arXiv:2212.05171}, 2022.

\bibitem{xue2023ulip}
Le Xue, Ning Yu, Shu Zhang, Junnan Li, Roberto Mart{\'\i}n-Mart{\'\i}n, Jiajun
  Wu, Caiming Xiong, Ran Xu, Juan~Carlos Niebles, and Silvio Savarese.
\newblock Ulip-2: Towards scalable multimodal pre-training for 3d
  understanding.
\newblock {\em arXiv preprint arXiv:2305.08275}, 2023.

\bibitem{zhao2021point}
Hengshuang Zhao, Li Jiang, Jiaya Jia, Philip~HS Torr, and Vladlen Koltun.
\newblock Point transformer.
\newblock In {\em Proceedings of the IEEE/CVF international conference on
  computer vision}, pages 16259--16268, 2021.

\end{thebibliography}
}

\end{document}